# Computer vision for liquid samples in hospitals and medical labs using hierarchical image segmentation and relations prediction

Sagi Eppel[*,1,2], Haoping Xu[2,3], Alan Aspuru-Guzik[*,1,2,3,4]

## Abstract

This work explores the use of computer vision for image segmentation and classification of medical fluid samples in transparent containers (for example, tubes, syringes, infusion bags). Handling fluids such as infusion-fluids, blood, and urine samples is a significant part of the work carried out in medical labs and hospitals. The ability to accurately identify and segment the liquids and the vessels that contain them from images can help in automating such processes. Modern computer vision typically involves training deep neural nets on large datasets of annotated images. This work presents a new dataset containing 1,300 annotated images of medical samples involving vessels containing liquids and solid material. The images are annotated with the type of liquid (e.g., blood, urine), the phase of the material (e.g., liquid, solid, foam, suspension), the type of vessel (e.g., syringe, tube, cup, infusion bottle/bag), and the properties of the vessel (transparent, opaque). In addition, vessel parts such as corks, labels, spikes, and valves are annotated. Relations and hierarchies between vessels and materials are also annotated, such as which vessel contains which material or which vessels are linked or contain each other. Three neural networks are trained on the dataset: One network learns to detect vessels, a second net detects the materials and parts inside each vessel, and a third net identifies relationships and connectivity between vessels. Dataset is available at this URL. Code is available at this URL.

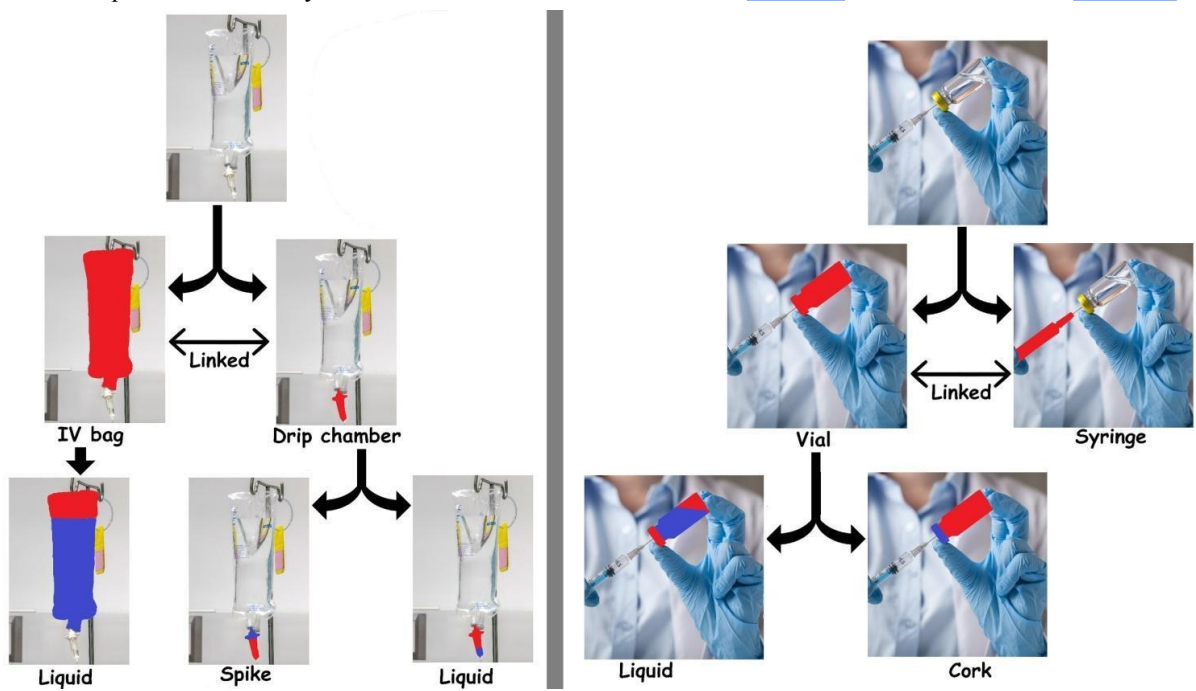

**Figure 1. Hierarchical segmentation of images into vessels and their content. The first layer segments the image into vessels. Each vessel is segmented into the liquids and parts it contains. The relationships and connectivity between the vessels and materials are also annotated.**

[1]Department of Chemistry, [2]Vector Institute, [3]Department of Computer Science University of Toronto,
[4]CIFAR Lebovic Fellow. Emails: sagieppel@gmail.com, haoping.xu@mail.utoronto.ca, alan@aspuru.com

## 1. Introduction:

The outbreak of the COVID-19 pandemic has overwhelmed many hospitals and medical laboratories and has highlighted the need to dramatically increase testing rates and reduce the workload of medical staff [1-3]. A considerable proportion of the work carried out in medical labs and hospitals involves handling fluid samples for various purposes, including blood and urine tests, infusion bags, and IV fluids (Figure 1). It is clear that automation of these tasks is essential in order to accelerate testing rates and to increase the capacity of medical facilities. Although a significant amount of work has been done on the automation of these systems[4-15], the main limitation of many automated systems is the reliance on human vision for many tasks involving fluid handling. Without using vision, it is unlikely that robotic systems will be able to handle many of the tasks currently done by humans. So far, most published works related to the use of vision in such systems have applied classic computer vision algorithms such as edge detectors for the detection of liquid level[16-19]. While these methods can work in controlled environments and with simple systems, they are usually insufficient for real-world scenarios, which tend to have high variability in terms of the environment, materials, and light. In recent years, the focus of computer vision has switched from handcrafted algorithms to machine learning approaches based on deep neural networks[20-22]. Deep learning uses large datasets of annotated images to train neural networks to detect the content of an image. This approach allows a neural net to learn to recognize various properties under a broad range of conditions, often achieving or surpassing human levels of accuracy. The main limitation of this approach is the need to collect and annotate a large number of images[20-22].

### 1.1. LabPics medical dataset

In this work, we introduce the LabPics Medical dataset, which contains 1,300 images taken from medical labs, hospitals, and clinics under a wide range of conditions. The images are annotated based on both instance and semantic segmentation of the liquids, materials, parts, and vessels (Figure 2). Semantic segmentation[23] involves finding image regions belonging to specific classes (e.g., Liquid, Solid), while instance segmentation[23] involves finding regions of different material phases and objects in the image (Figure 2). The phase of the material (e.g., liquid, foam, suspension, or solid) and the type of liquid (blood, urine) are annotated (Figure 2), as are the types of vessel (e.g., syringes, cups, tubes, infusion bags, drip chambers), and its properties such transparency or opacity. Relationships and hierarchies (e.g., which vessel contains which material or other vessels, or which vessel pairs are connected) are also annotated (Figure 1). The vessel parts such as corks, labels, and spikes are also annotated.

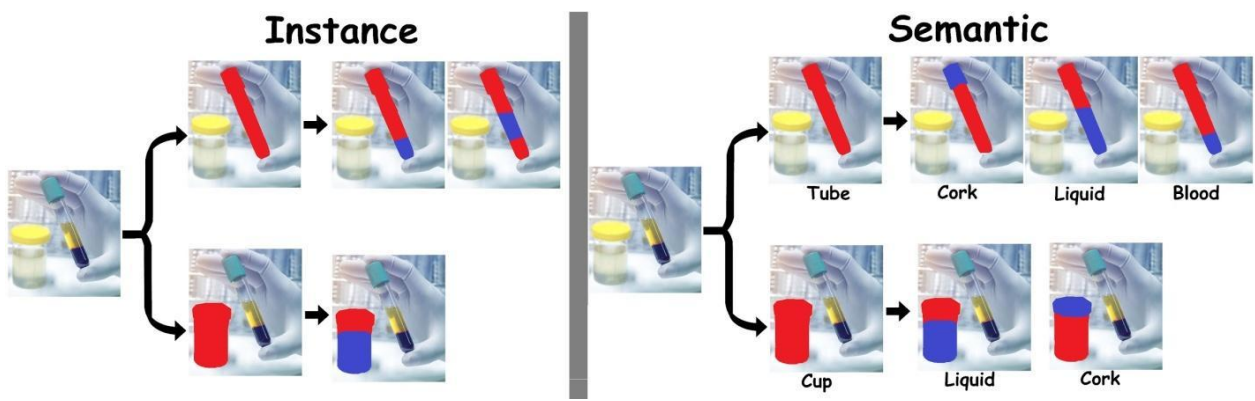

**Figure 2. Hierarchical semantic and instance segmentation in the LabPics Medical dataset.** Each image is segmented into vessel instances (red), and the region of the vessel is further segmented into instances of liquid and other material phases (Left, blue). Semantic maps are also provided for materials and vessel parts in which all the pixels in each vessel belonging to a given class are marked (right, blue).

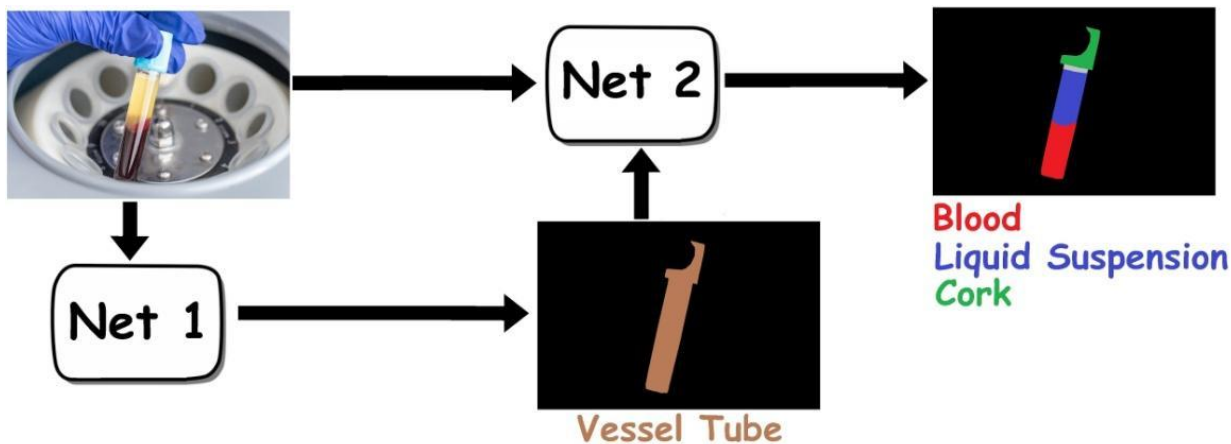

**Figure 3.** Modular hierarchical instance segmentation. Net 1 segments the vessel instance from the image. Net 2 receives the vessel instance mask (from Net 1) and the image and identifies the materials and parts inside the vessel.

A main limitation of the dataset is its small size. However, the problem of handling medical fluids is similar to the problem of handling materials in vessels in a chemistry lab. This problem is already dealt with in the previously published LabPics chemistry dataset[24]. We are releasing a beta version of a new LabPics Chemistry dataset containing 6,500 images from chemistry lab experiments, which are annotated in the same format as the medical dataset and used to support training. For more details, see section 4. The dataset is available from [this URL](#).

### 1.2. Hierarchical image segmentation

We use our dataset to demonstrate a new set of cascade modular neural networks[25-29] that combine hierarchical semantic and instance segmentation and can classify both the vessels and the materials inside them. One neural network is used to detect and segment the vessels (Figure 3, Net 1), while a separate neural net receives the vessel region produced by the first net and segments the materials and parts inside this vessel (Figure 3, Net 2). The two nets are trained independently but operate together via a hierarchical cascading scheme (Figure 3). This allows the system to identify not only instances of vessels and materials but also their hierarchies, meaning which liquid belongs to which vessel. For example, if a mask of a pipette containing blood is used as the input Net 2 (Figure 3), the net will output the segment representing the blood (Figure 4, left); however, if the input to the net is a mask of a tube containing a pipette, the net will not predict the blood region (Figure 4, right). While the blood is inside the tube region, the net learns the hierarchies between the vessels and identifies the blood is not directly inside the tube but is inside the pipette (Figure 4). For vessel detection (Figure 3, Net 1), we use the Mask RCNN[30] object detection net. While for the content detection (Figure 3, Net 2) we used a Fully convolutional net (FCN)[32-33] trained for combined instance and semantic segmentation (Section 6).

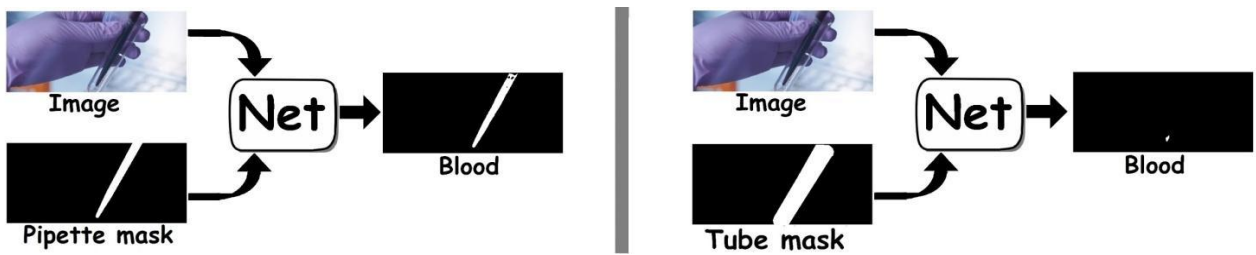

**Figure 4:** Left: If the pipette mask is used as an input for the net, the net predicts the blood segment directly inside the pipette; right: If the mask of the tube containing the pipette is used as input, the net will ignore the blood inside the pipette and will only predict the direct content of the tube.

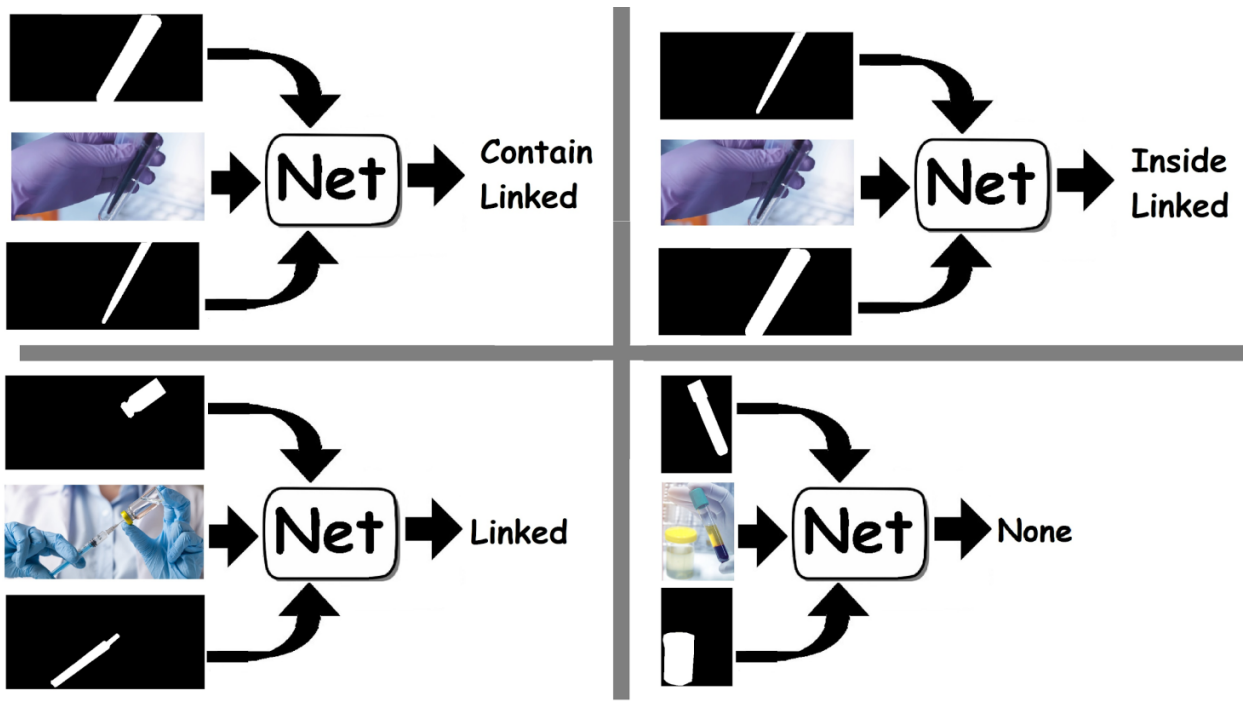

**Figure 5. Predicting relationships between vessels.** The net receives the image and masks of the two vessels and predicts the relationship. A 'linked' relationship means that there is a connection that allows liquid to pass between the two vessels, while 'inside' or 'contain' means that one vessel contains the other.

### 1.3. Relationship prediction

The relationships between vessels represent another crucial aspect of understanding and handling liquids in medical systems and other fields[34-35] (Figure 5). Understanding which vessels are inside or linked to other vessels is essential when transferring liquids between vessels or when manipulating and moving vessels (Figure 5). For example, in the case of an IV-bag, it is essential to understand which drip-chamber it is linked to (Figure 5). We treat the relations predictions as a classification problem with three nonexclusive classes for the relationships between vessel pairs: 'inside' or 'contain' if one vessel is inside another, and 'linked' if there is a connection that allows fluid to pass between the vessels (Figure 5). Each of these relations may exist or be absent independently of the others. To predict these relations, we use a classification net that receives the image and the masks of the two vessels and classifies the relationship between them (Figure 5). Details on the architecture and training of this net can be found in section 5.

## 2. Evaluation metrics

For evaluating the prediction accuracy of the nets on the segmentation of classification tasks, we use two standard metrics of intersection over union (IOU)[32] for semantic-segmentation and panoptic quality (PQ)[36], for instance-segmentation.

### 2.1. Intersection over union (*IOU*) for semantic segmentation evaluation

The intersection over union (*IOU*) is the main metric used to evaluate semantic segmentation and is calculated separately for each class. The intersection is the sum of the pixels that belong to the class according to both the net prediction and the dataset ground truth (GT), while the union is the sum of pixels that belong to the class based on either the net prediction or the GT. *IOU* is the intersection divided by the union. In addition, the recall is the intersection divided by the sum of all pixels belonging to the class according to the GT annotation, while precision is the intersection divided by the sum of all pixels belonging to the class based on the net prediction.

## 2.2. Panoptic quality (PQ)

For instance segmentation, we choose to use the standard metric of panoptic quality (*PQ*). This is a combination of the recognition quality (*RQ*) and segmentation quality (*SQ*), where a segment is defined as the region of each individual instance in the image (instances are material phases or vessels). *RQ* is used to measure the detection rate of instances, and is given by $RQ = \frac{TP}{TP+(FP+FN)\times 0.5}$, where *TP* (true positive) is the number of predicted segments that match a GT segment; *FN* (false negative) is the number of GT segments that do not match any of the predicted segments, and *FP* (false positive) is the number of predicted segments with no matched GT segment. Matching is defined as an *IOU* of 50% or more between predicted and GT segments of the same class. *SQ* is simply the average *IOU* of matching segments, and *PQ* is calculated as:
$PQ = RQ \times SQ$.

## 2.3. Class-agnostic *PQ* metrics

The standard *PQ* metric is calculated by considering only those segments that are correctly classified. This means that if a predicted segment overlaps with a GT segment but has a different class, it will be considered mismatched. The problem with this approach is that it does not measure the accuracy of segmentation without classification; a net that predicts the segment region perfectly but with the wrong class will have a *PQ* value of zero. One method of overcoming this problem is to pretend that all segments have the same class: in this case, the *PQ* will depend only on the region of the predicted segment. However, given the class imbalance, this will increase the weight of the more common classes, and will not accurately measure the segmentation accuracy across all classes. To measure class-agnostic segmentation in a way that will equally represent different classes, we use a modified *PQ* metric. The *PQ, RQ,* and *SQ* values for the class-agnostic method are calculated in the same way as in the standard case, while the definitions of *TP, FP*, and *FN* are modified. In this case, the *TP* for a given class is the number of GT instances of this class that match predicted instances with IOU > 0.5 (regardless of the predicted instance class). The *FN* for a given class is the number of GT instances of this class that does not match any predicted segment (regardless of the predicted segment class). If an instance has more than one class, it will be counted separately for each class. The *FP* for a given class is the fraction of GT segments that belong to this class multiplied by the total number of class-agnostic *FP* segments. The total number of class-agnostic *FP* segments is the number of predicted segments that do not match any GT segments, regardless of class (where matching means IOU > 0.5 between segments, regardless of class). For example, if 20% of the GT segments belong to the solid class, and there are 1,200 predicted segments that do not match any GT segments, the *FP* for the solid class would be 1,200 × 0.2 = 240. In other words, to avoid using the predicted class for the *FP* calculation, we split the total *FP* among all classes based on the class ratio in the GT annotation.

## 2.4. Region-specific evaluation of vessel content

Segmentation of the vessel content was evaluated separately for each vessel, and the results were averaged. Two settings were used to predict the vessel content: the first used the vessel region from the GT dataset as an input to the material prediction net (Figure 3, Net 2), while in the second, the vessel region predicted by the Mask RCNN at the vessel detection stage was used as the input to the material prediction net (Figure 3, Net 1).

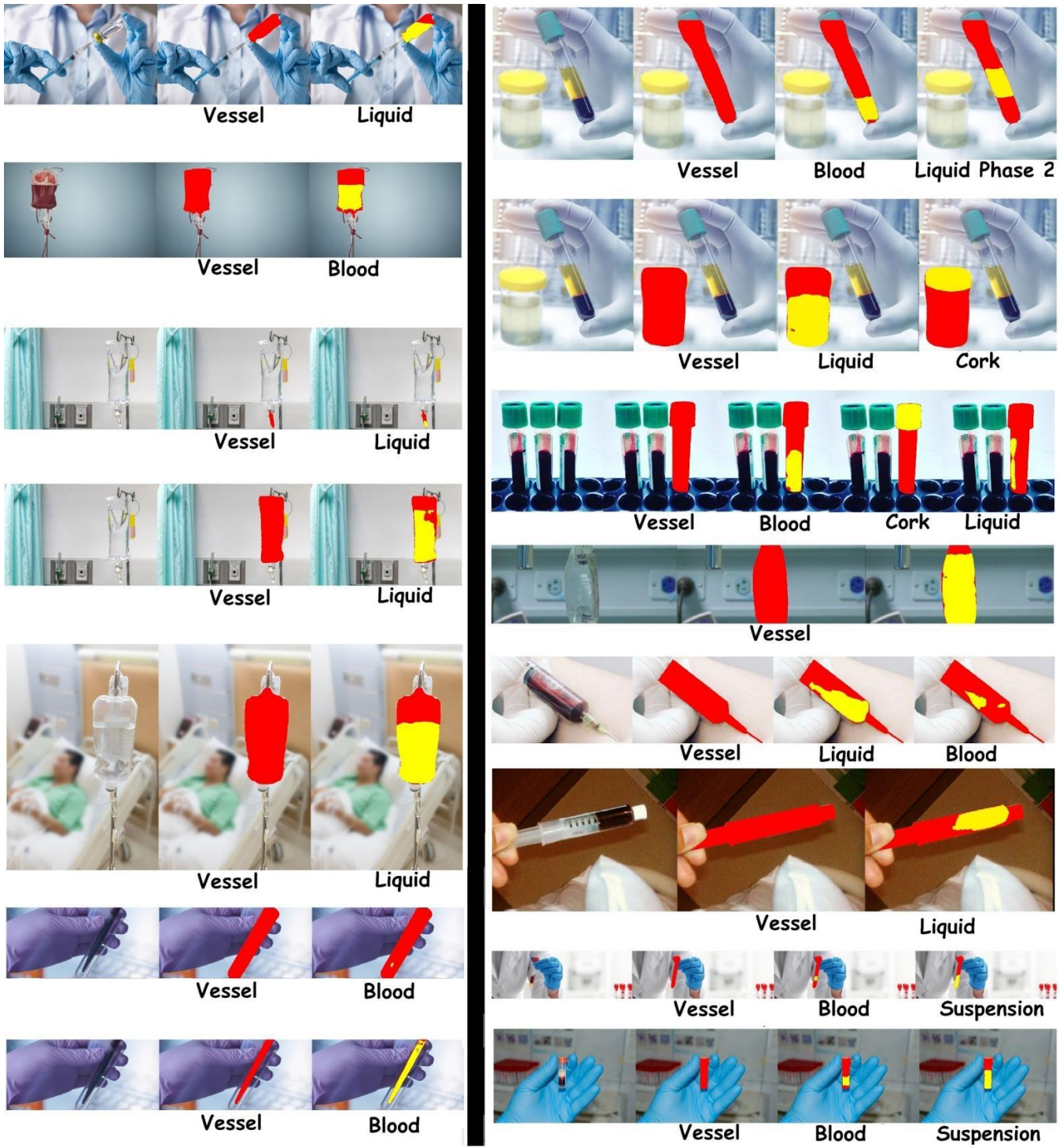

**Figure 6. Selected results from the vessel content prediction nets. The vessel mask that was used as an input to the net is marked in red. Each panel shows a predicted segment inside the vessel (yellow) and its predicted class below the panel.**

## 3. Results

### 3.1. Instance segmentation of vessel content, liquids, materials, and parts

The results of the material prediction net (Figure 3, Net 2) are given in Table 1, with selected results shown in Figure 6. It can be seen from the class-agnostic PQ values in Table 1 that the net achieves good accuracy in terms of detecting and segmenting for all classes of material except foam. These results are consistent across a wide range of vessels and environments (Figure 6). Foam is the only class with poor segmentation accuracy (Table 1). This can be explained by the fact that foam tends to form a narrow band at the top of a liquid or to appear in small, sparse regions; both options are hard to segment. When Evaluating segmentation with classes (Table 1), the accuracy differs significantly between the more common classes, such as liquids, and rarer classes, such as gel or blood. The classification and segmentation accuracy for abundant classes like liquids remain high (Table 1), while for rarer classes such as blood and solids, the accuracy is lower than the class-agnostic case by about 20% on average.

#### 3.1.1. Effect of the vessel mask source

The system was tested (Figure 3) using two settings: in the first, the vessel mask used as input to the net (Figure 3, Net 2) was taken from the dataset Ground Truth (GT) annotation. In the second, the vessel mask used as input was predicted by MaskR-CNN neural net. Both settings used the exact same net for material detection (Figure 3, Net 2). This net was trained with only the GT vessel mask as input (Section 6.8). When testing the net, the results of the system that used the GT mask as input were slightly better. However, the difference was not significant (Table 1). Hence, the net was able to effectively use the imperfect predicted vessel mask as input, despite being trained only on perfect GT masks. This demonstrates that modular nets are flexible and allow replacing one module (in this case, the mask detector) by others without retraining and with no significant accuracy reduction.

#### 3.1.2. Predicting hierarchies

It can be seen from the two examples in Figure 4 that the net learned to understand hierarchies. When the input was the mask of a pipette, the net predicted the segment of blood inside this pipette (Figure 4, Left). However, when the input was a tube containing a pipette, the net ignored the blood inside the pipette, even though it was inside the vessel mask region (Figure 4, Right). This implies that the net learned to recognize that the blood was in the pipette rather than directly in the tube. Hence the net learned to understand context and hierarchies.

#### 3.1.3. Semantic segmentation of vessel content

The semantic segmentation IOU results for common classes such as liquids, suspension, solids, labels, and corks were significantly better than for the rarer classes (e.g., urine, blood), which yielded poor accuracy (Table 2). Semantic segmentation was applied using two methods: in the hierarchical approach, the vessel mask produced by the Mask-RCNN was used as input, and a semantic map was predicted for each vessel separately (Figure 3, Net 2), while in the second approach, a standard FCN net was used to predict the full image segmentation map. The hierarchical approach gave far better accuracy (Table 2). This can be attributed to the use of the state-of-the-art Mask R-CNN in the vessel detection stage.

### 3.2. Prediction of vessel relationships

Results for the prediction of the relationships between pairs of vessels (Figure 5) are given in Table 3. It can be seen that good accuracy was achieved for all relationship classes. Pairs of vessels with no connectivity or relationship were by far the most common, and this relationship was predicted with perfect accuracy ('None', Table 3). The other relationships were also predicted with good accuracy (Table 3). One unexpected finding was that the precisions for the 'inside' and 'contain' relationships were not identical (Table 3), although the examples for both cases were identical and depended only on the order in which the vessels were introduced to the net (Figure 5).

Table 1: Instance segmentation results for liquid and material phases inside the vessel

| | GT vessel [a] | | | | | | Hierarchical Mask RCNN for vessel [b] | | | | | |
| --- | --- | --- | --- | --- | --- | --- | --- | --- | --- | --- | --- | --- |
| | Class-agnostic [c] | | | With class [d] | | | Class-agnostic [c] | | | With class [d] | | |
| Class | PQ | SQ | RQ | PQ | SQ | RQ | PQ | SQ | RQ | PQ | SQ | RQ |
| Filled | 0.59 | 0.81 | 0.73 | 0.59 | 0.81 | 0.73 | 0.53 | 0.79 | 0.66 | 0.53 | 0.79 | 0.66 |
| Liquid | 0.60 | 0.81 | 0.74 | 0.60 | 0.81 | 0.74 | 0.53 | 0.80 | 0.67 | 0.53 | 0.80 | 0.66 |
| Suspension | 0.58 | 0.80 | 0.73 | 0.51 | 0.80 | 0.73 | 0.50 | 0.78 | 0.65 | 0.43 | 0.78 | 0.55 |
| Blood | 0.62 | 0.80 | 0.77 | 0.45 | 0.80 | 0.77 | 0.53 | 0.79 | 0.67 | 0.37 | 0.82 | 0.45 |
| Urine | 0.70 | 0.83 | 0.84 | 0.23 | 0.83 | 0.84 | 0.65 | 0.81 | 0.80 | 0.22 | 0.82 | 0.27 |
| Solid | 0.52 | 0.65 | 0.81 | 0.40 | 0.65 | 0.81 | 0.51 | 0.64 | 0.79 | 0.44 | 0.66 | 0.67 |
| Powder | 0.54 | 0.69 | 0.79 | 0.34 | 0.69 | 0.79 | 0.50 | 0.64 | 0.78 | 0.23 | 0.57 | 0.40 |
| Granular | 0.47 | 0.51 | 0.93 | 0.34 | 0.51 | 0.93 | 0.53 | 0.58 | 0.92 | 0.39 | 0.58 | 0.67 |
| Foam | 0.21 | 0.64 | 0.32 | 0.20 | 0.64 | 0.32 | 0.20 | 0.64 | 0.32 | 0.20 | 0.64 | 0.32 |
| Gel | 0.89 | 0.96 | 0.93 | 0.00 | 0.96 | 0.93 | 0.88 | 0.96 | 0.92 | 0.00 | 0.00 | 0.00 |

a) The GT mask from the dataset was used as input to the net (Figure 3, Net 2).
b) The vessel mask predicted by the Mask R-CNN was used as input to the net (Figure 3, Net 2).
c) Matching between the GT and predicted segments depended only on segment overlap, and not on class (Section 2.3).
d) Standard PQ metrics: matching GT and predicted segments must have the same class (Section 2.2).

Table 2: Results of semantic segmentation

| | Hierarchical GT vessel [a] | | | Hierarchical Mask RCNN vessel [b] | | | Full image [c] | | |
| --- | --- | --- | --- | --- | --- | --- | --- | --- | --- |
| | mIOU [d] | Precision | Recall | mIOU | Precision | Recall | mIOU | Precision | Recall |
| Filled | 0.69 | 0.91 | 0.74 | 0.65 | 0.92 | 0.69 | 0.59 | 0.87 | 0.65 |
| Liquid | 0.65 | 0.88 | 0.72 | 0.62 | 0.88 | 0.68 | 0.58 | 0.85 | 0.65 |
| Suspension | 0.50 | 0.81 | 0.56 | 0.47 | 0.79 | 0.54 | 0.41 | 0.81 | 0.45 |
| Blood | 0.27 | 0.94 | 0.28 | 0.26 | 0.97 | 0.26 | 0.20 | 0.91 | 0.21 |
| Urine | 0.21 | 0.96 | 0.21 | 0.20 | 0.96 | 0.20 | 0.06 | 0.71 | 0.07 |
| Solid | 0.40 | 0.47 | 0.73 | 0.40 | 0.48 | 0.72 | 0.25 | 0.30 | 0.58 |
| Powder | 0.31 | 0.59 | 0.40 | 0.30 | 0.66 | 0.36 | 0.24 | 0.34 | 0.46 |
| Granular | 0.19 | 0.21 | 0.70 | 0.20 | 0.22 | 0.68 | 0.23 | 0.24 | 0.86 |
| Foam | 0.21 | 0.69 | 0.24 | 0.20 | 0.66 | 0.22 | 0.13 | 0.75 | 0.14 |
| Gel | 0.02 | 0.39 | 0.02 | 0.01 | 0.29 | 0.01 | 0.17 | 0.33 | 0.26 |
| Part | 0.59 | 0.95 | 0.61 | 0.55 | 0.96 | 0.56 | 0.57 | 0.82 | 0.65 |
| Label | 0.48 | 0.98 | 0.48 | 0.48 | 0.98 | 0.48 | 0.57 | 0.90 | 0.61 |
| Spike | 0.51 | 0.92 | 0.54 | 0.43 | 0.92 | 0.45 | 0.34 | 0.94 | 0.35 |
| Cork | 0.65 | 0.93 | 0.68 | 0.59 | 0.93 | 0.61 | 0.59 | 0.91 | 0.63 |
| Opaque | 0.17 | 1.00 | 0.17 | 0.17 | 0.99 | 0.17 | 0.49 | 0.79 | 0.56 |
| Transparent | 0.88 | 0.92 | 0.95 | 0.82 | 0.93 | 0.87 | 0.74 | 0.86 | 0.84 |

a) The GT mask from the dataset was used as input to the net (Figure 3, Net 2).
b) The vessel mask predicted by the Mask R-CNN was used as input to the net.
c) Prediction on the entire image using standard FCN rather than hierarchical (Section 6.9).
d) Intersection over union (Section 2.1).

Table 3: Results for the prediction of relationships between vessels (Section 1.3)

| Relationship | Precision [a] | Recall [b] | mIOU [c] |
| --- | --- | --- | --- |
| Linked | 95% | 85% | 82% |
| Inside | 88% | 83% | 75% |
| Contain | 94% | 83% | 79% |
| None | 99% | 99% | 99% |

a) Precision=TP/(TP+FP). TP is the true positive rate, FP is the false positive rate.
b) Recall=TP/(TP+FN). FN is the false-negative rate.
c) IOU=TP/(TP+FP+FN)

### 3.3. Segmentation of vessel instances

The results for vessel segmentation are shown in Table 4 and in Figure 7. It can be seen that the Mask R-CNN gave good results for detection and segmentation (without class) for most vessels (Table 4, class agnostic). When including classification errors, the Mask-RNN gave poor accuracy for most classes (Table 4, with class). This can be explained by the fact that the Mask R-CNN was pre-trained for object detection and could easily generalize to the segmentation of new vessel objects. However, classification is a less transferable task than segmentation, and in this case, the net had to rely only on the limited number of examples in the LabPics Medical dataset.

**Table 4: Instance segmentation results for vessels**

|  | Class-agnostic [a] | | | With class [b] | | |
|---|---|---|---|---|---|---|
|  | PQ | SQ | RQ | PQ | SQ | RQ |
| Vessel | 0.73 | 0.86 | 0.85 | 0.73 | 0.86 | 0.85 |
| Tube | 0.72 | 0.85 | 0.84 | 0.54 | 0.87 | 0.62 |
| IV bag | 0.77 | 0.87 | 0.89 | 0.41 | 0.89 | 0.46 |
| IV bottle | 0.58 | 0.81 | 0.71 | 0.00 | 0.00 | 0.00 |
| Drip chamber | 0.70 | 0.81 | 0.86 | 0.00 | 0.00 | 0.00 |
| Bottle | 0.65 | 0.85 | 0.77 | 0.00 | 0.00 | 0.00 |
| Syringe | 0.64 | 0.78 | 0.81 | 0.00 | 0.00 | 0.00 |
| Pipette | 0.34 | 0.67 | 0.51 | 0.00 | 0.00 | 0.00 |
| Beaker | 0.83 | 0.88 | 0.95 | 0.00 | 0.00 | 0.00 |
| Bottle | 0.65 | 0.85 | 0.77 | 0.00 | 0.00 | 0.00 |
| Bowl | 0.84 | 0.89 | 0.95 | 0.00 | 0.00 | 0.00 |
| Cup | 0.89 | 0.93 | 0.95 | 0.30 | 0.94 | 0.32 |
| Plate | 0.92 | 0.96 | 0.95 | 0.00 | 0.00 | 0.00 |
| Flask | 0.67 | 0.86 | 0.78 | 0.00 | 0.00 | 0.00 |
| Jar | 0.90 | 0.95 | 0.95 | 0.00 | 0.00 | 0.00 |
| Transparent | 0.73 | 0.86 | 0.85 | 0.00 | 0.00 | 0.00 |
| Opaque | 0.92 | 0.96 | 0.95 | 0.00 | 0.00 | 0.00 |
| Vessel inside vessel | 0.45 | 0.72 | 0.62 | 0.00 | 0.00 | 0.00 |

a) Matching between GT and predicted segments depended only on segment overlap, and not on class (Section 2.3).
b) Standard PQ metrics. Matching GT and predicted segments must have the same class (Section 2.2).

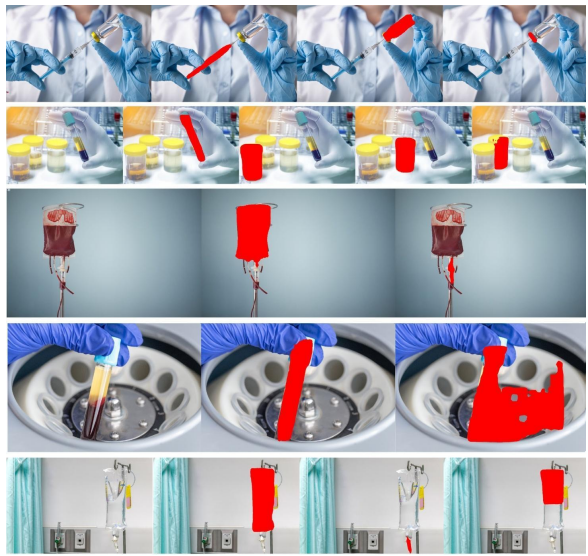

Figure 7. Vessel predicted using Mask-RCNN. Each panel shows a predicted vessel instance, marked in red.

# Appendix

## 4. Preparation and structure of the LabPics Medical dataset

The goal of the LabPics medical dataset is to reflect the different ways in which liquid samples are handled in medical labs, hospitals, and other fields related to health care. The images in the dataset were publically shared by various open sources on the internet, mainly from YouTube and Instagram. The images include infusion and blood bags, drip chambers, syringes, IV fluids, blood samples, and urine samples in various systems, as well as other medical fluid samples. The images were manually annotated using the VGG image annotation tool (VIA)[37]. If several vessels are connected together, each one is considered to be a separate instance (Figure 1). If a vessel contains several materials in different phases, such as phase-separating liquids, a solid immersed in a liquid, or a liquid with foam, each phase is marked separately (Figure 2). The instances can overlap: a solid may be immersed in a liquid, materials and other parts may be inside vessels, and vessels may be inside other vessels. For each instance, there is a list of classes to which it belongs. Instances can belong to several classes and usually have both a class and a subclass. For example, instances belonging to the 'tube' class will also belong to the 'vessel' class, while instances that belong to the 'blood' class will also belong to the 'liquid' class. Another set of classes is used to define the properties of the instance. For vessels, the properties are 'transparent', 'semi-transparent, and 'opaque'. For materials, the properties include 'scattered,' representing materials that do not fill the volume but are scattered in the form of sparse particles, and 'on-surface,' which is used for materials that do not fill the volume but form a layer on the surface of the vessel. Finally, the relationships between instances are annotated. For each instance, other instances that contain it or inside it are annotated (Figures 4). This includes materials inside vessels, vessels inside vessels, or materials immersed inside other materials (such as solids immersed in liquids). In addition, the connectivity of the vessels is annotated; for example, an instance of an infusion bag may be connected to a drip chamber, or a syringe may be connected to a vial (Figure 1).

### 4.1. LabPics Chemistry

We have also released a beta version of the LabPics Chemistry dataset, containing 6,500 images from chemistry labs that are annotated using the same format as the LabPics Medical dataset, in order to support training on LabPics Medical and to compensate for its small size.

### 4.2. Additional datasets

Although there are no other datasets that are similar to LabPics, some properties of this dataset are shared by other more general datasets. The Trans10k dataset contains semantic segmentation for transparent objects in a large number of scenes[38]. Transparency is one of the main property classes in the LabPics dataset and is also the main feature of many of the vessels. In addition, the COCO dataset[21,35] contains many vessel objects, such as glasses, bottles, cups, and bowls, in a large number of scenes. Combining these two datasets with the LabPics dataset in the training stage allows the nets to become exposed to a large number of new environments and objects. Although this does not improve the accuracy of the net on the LabPics test set, it considerably improves the robustness of the nets in a complex environment in which there are many objects besides a vessel.

### 4.3. Training and test sets

The dataset was divided into two subsets: a training set containing 1,000 images and a test set containing 300 images. The images in both sets were chosen from different YouTube and Instagram channels in order to maximize the differences in the settings and environments between the training and testing.

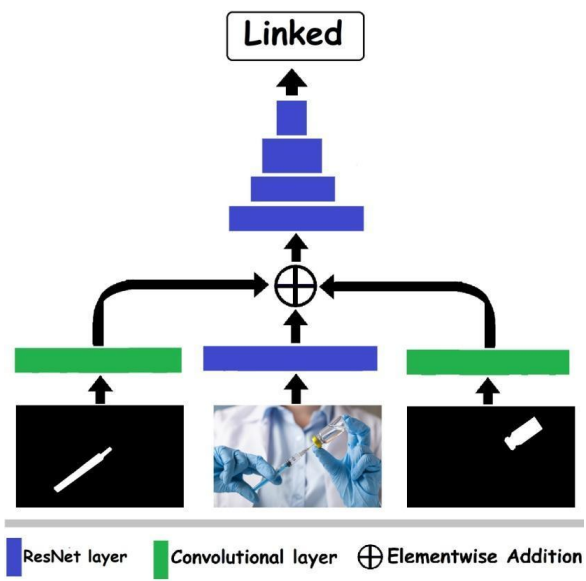

Figure 8. Predicting the relationships between two instance classes. The prediction was generated using a modified ResNet for image classification, with the addition of the two vessel masks as inputs. Each vessel mask was processed using a separate convolution layer, and the output of these layers was added to the feature map after the first ResNet layer.

## 5. Structure of vessel relationships prediction net

The relationships between two vessels (Figure 5) were identified using modified neural nets for image classification (ResNet[39]). The image and the masks of the two vessels were used as inputs to the net (Figure 8). The vessel masks were processed using a separate convolutional layer (a single layer for each mask), and the output feature maps were added (elementwise) to the feature map produced by the image after the first ResNet layer (Figure 8). The relationships were limited to three classes: 'inside', 'contain', and 'linked'. Each class was predicted independently in the form of two values, which after normalization using Softmax give the probability that the class exists or not. The 'inside' and 'contain' relations were interchangeable: replacing the order of the masks changed the prediction from 'inside' to 'contain', and vice versa (Figure 5). The training was carried out on each pair of vessels in each image. The choice of pairs for each training batch was made randomly with equal probability for each relationship class (to ensure class balance).

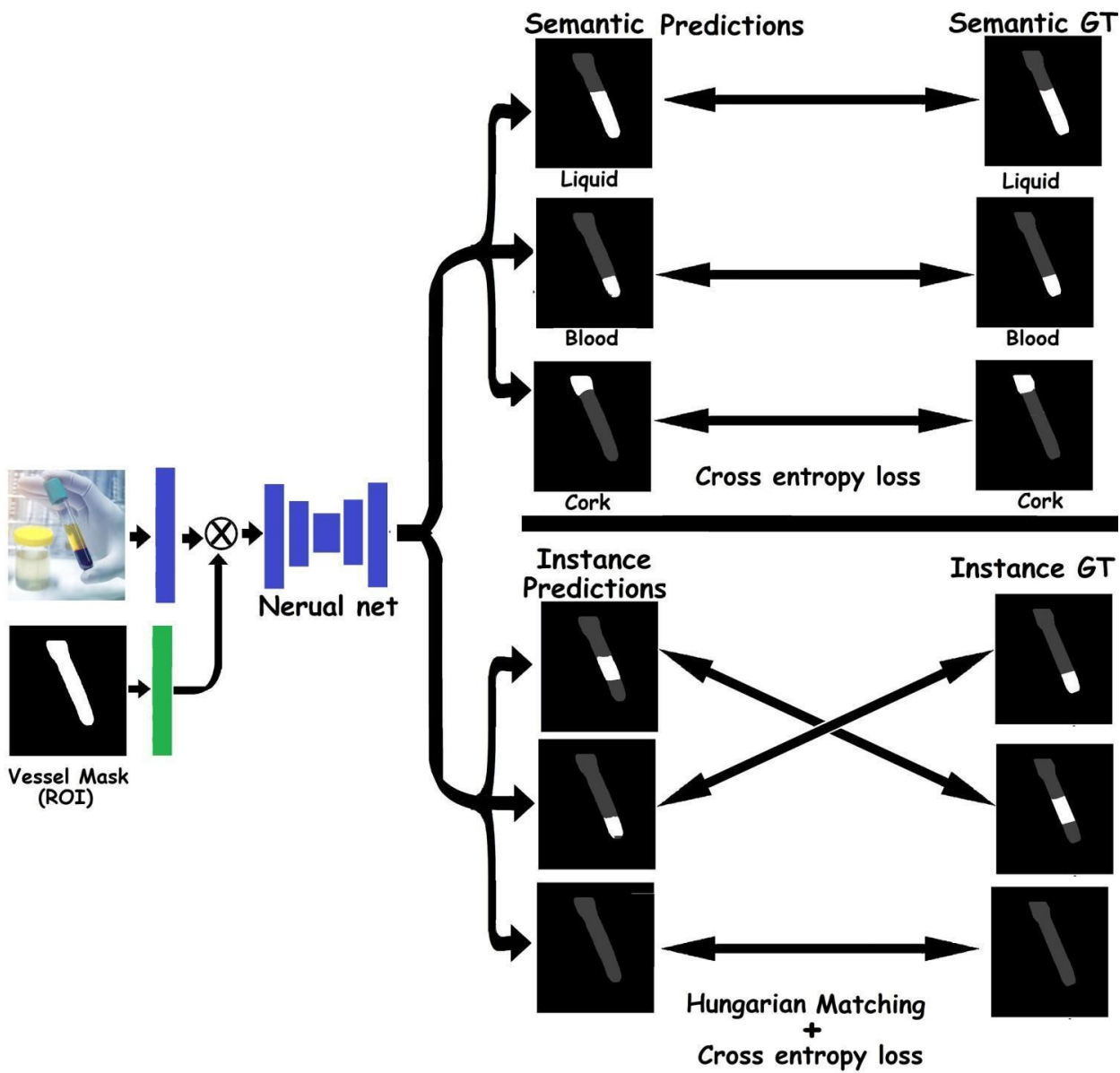

Figure 9. Structure of a neural net for predicting the content of a vessel. The net is based on a fully convolutional neural net (FCN) with the vessel mask as input. The vessel mask is processed using a single convolutional layer and is merged with the first feature map of the main net by elementwise multiplication. The output of the net is a set of binary masks for each segment. Predicted segment masks are divided into semantic masks and instance masks: Predicted semantic masks are matched to their corresponding GT semantic masks based on class. Predicted instance masks are matched to the closest instance in the GT masks (based on the IOU similarity) using the Hungarian matching algorithm. The loss for each mask is calculated using the elementwise cross-entropy function between GT and the predicted mask. Note that the vessel mask is overly on the output masks (dark grey) for visualization only and is not really part of the output.

# 6. Architecture and training of the nets

## 6.1. Hierarchical instance segmentation using modular nets

Identifying materials in vessels can be achieved either by identifying both the vessel and its content in one step, using a single system or in a hierarchical way, by first identifying the vessel using one system and then using the known vessel region to identify its content. Identifying first the vessel and then its content is a more intuitive approach and considerably simplifies the problem; it has also been shown to give better results[24-29], as it allows the problems of vessel detection and content segmentation to be dealt with and evaluated separately. This creates a much better understanding of the system's performance and makes the optimization of each task easier. In addition, this modular approach allows each module to be trained or replaced without the need to retrain the overall system, meaning that each network can be used for other purposes or replaced by another system. In many robotic systems, the vessel region is either static or known in advance, meaning that the system does not need to continuously recognize the vessel region and only needs to track its content. This can easily be done in a modular system by replacing the vessel detection net with a known vessel region.

## 6.2. Predicting vessel content, materials, and parts using region-specific segmentation

Detecting and segmenting the vessel content is achieved using a fully convolutional neural net (FCN), with a vessel region mask as an additional input (Figure 9). This net segments only the content of the vessel passed to it as input. The net receives both an image and a vessel mask as input, and outputs masks corresponding to the different segments of materials and parts inside the vessel (Figure 9). These segments are divided into two groups: semantic maps, in which each mask corresponds to a given class, and instance masks, in which each mask corresponds to a single material phase (Figure 9).

## 6.3. Predicting semantic maps

For each class (liquid, solid, blood, label, cork, etc.), the net will produce two output maps: one for the probability that a pixel belongs to the class and another for the probability that the pixel does not belong to the class. Normalization of each mask pair using Softmax gives the probability the pixel belongs to a given class.

## 6.4. Predicting instance maps

Instance masks are again predicted using two maps: one for the probability that a pixel belongs to the instance and a second for the probability that it does not belong to the instance. One problem is that the number of instances in the image may change in each case, but the number of predicted masks is constant. To solve this problem, the net always predicts ten instance masks. The predicted masks may be empty if less than ten instances are available (Figure 9); for example, if the vessel contains two material instances, eight of the predicted masks will be empty (Figure 9). This was based on the assumption that the occurrence of more than ten material phases in a single vessel is very unlikely.

## 6.5. Loss function

The loss for the semantic maps was simply the cross-entropy between predicted and the GT semantic mask of the same class (Figure 9). The loss for the instance maps is calculated by first matching a single GT instance for each predicted instance mask (Figure 9). Matching is done by first calculating the intersection over union (IOU) between each predicted instance mask and each GT mask, and then using the Hungarian matching algorithm to assign one GT instance to each predicted instance[40,41] (Figure 9). The number of GT instance masks is always matched to the number of predicted masks by adding empty masks (Figure 9). Once the assignment has been made, the loss is again calculated based on the elementwise cross-entropy between each predicted mask and the corresponding GT masks (Figure 9).

### 6.6. Adding the vessel mask as input

A standard FCN receives only an image as input. Adding the vessel mask as a second input is achieved by using the vessel mask as an image and passing it through a single convolution layer (Figure 9). The output of this layer merged with the FCN first layer output using element-wise multiplication. The resulting map is fed to the next layer of the net (Figure 9).

### 6.7. Assigning classes to instances

Instance masks are predicted without a class (Figure 9). To find the instance classes, each instance mask is matched with the semantic maps predicted by the net. If more than 33% of the area of the instance mask overlaps with a given semantic mask, the semantic mask's class is assigned to the instance. This is done only for the material classes and may not work when several different instances overlap.

### 6.8. Architecture and training

The structure of the FCN (Figure 9) is based on DeepLab V3[42], with additional upsamples and skipped connection layers in a similar way to UNet[32]. The encoder for the FCN[31] is a pretrained ResNext50[39] followed by an ASPP[42] layer and three upsampling layers that use simple bilinear upsampling. At each training iteration, the loss for either the instances or semantic maps is used (but not both). The choice of which to use is made randomly, with a 50:50 probability. The classes used were all of the materials classes, part classes, material properties, and vessel properties. LabPics Chemistry was used in 65% of the iterations, and LabPics Medical in the remaining 35%. Again, the choice was random for each batch. The learning rate started at 1e-5 and was decreased by a factor of 0.9 after every 30,000 training steps, but only if the average loss was not decreasing during this period. If the learning rate went below 1e-6, it was increased to 5e-6. The training was carried out using the Adam optimizer on PyTorch with a Titan GTX GPU (16 GB).

### 6.9 Applying FCN to detect vessel instances and full image semantic maps

The approach described above can also be used to achieve full image segmentation by removing the vessel mask input. This approach is applied for full image semantic segmentation.

## 7. Vessel detection using Mask RCNN

The detection of vessels in an image is similar to other problems involving object detection and can be solved using existing instance segmentation methods. One of the leading object detection methods is the Mask R-CNN[30] neural net. Following previous work, our model uses ResNet as a backbone, followed by a region proposal network (RPN) that provides a list of candidate instances. For each candidate, both the instance bounding box and class are predicted by the box head. Masks for vessels are also generated. Since an instance in the dataset can belong to multiple subclasses, the original Mask-RCNN scheme is modified to handle multi-label classification. This function is enabled via an additional subclass predictor, which takes the same ROI feature generated by the box head and output label powerset as the multi-label subclass prediction. The predictor takes the same feature vector from the box head and uses a single fully connected layer to carry out the classification. The subclass loss is defined as a binary cross-entropy loss.

## 10. Acknowledgement

The dataset was annotated by Mor Bismuth. We acknowledge the Defense Advanced Research Projects Agency (DARPA) under the Accelerated Molecular Discovery Program under Cooperative Agreement HR00111920027, dated August 1, 2019. The content of the information presented in this work does not necessarily reflect the position or the policy of the Government. A.A.-G. thanks Anders G. Frøseth for his generous support. Most images used in this paper were taken with permission from Shutterstock.

## 11. Supporting material

The LabPics Dataset can be downloaded from:
https://zenodo.org/record/4736111
https://www.kaggle.com/sagieppel/labpics-chemistry-labpics-medical
Code and trained models are available at this URL.